# Approximation of Classification and Measures of Uncertainty in Rough Set on Two Universal Sets

B. K. Tripathy and D. P. Acharjya

*School of Computing Science and Engineering*
*VIT University, Vellore – 632014 (India)*
*tripathybk@rediffmail.com, dpacharjya@gmail.com*

*Abstract*

*The notion of rough set captures indiscernibility of elements in a set. But, in many real life situations, an information system establishes the relation between different universes. This gave the extension of rough set on single universal set to rough set on two universal sets. In this paper, we introduce approximation of classifications and measures of uncertainty basing upon rough set on two universal sets employing the knowledge due to binary relations.*

## 1. Introduction

Now-a-days Internet is the best example for distributed computing which involves dispersion of data geographically. Therefore in modern era of computing, for human being there is a need of development in data analysis from the huge amount of data available. Hence, it is very difficult to extract expert knowledge from the universe. Many new mathematical modeling tools such as fuzzy set [14], rough set [29, 30] and soft set [6] are emerging to the thrust of the real world task. Development of these techniques and tools and its popularity are studied under different domains like knowledge discovery in database, computational intelligence, knowledge engineering, granular computing etc. [17, 18, 19, 21, 22, 28].

The basic idea of rough set [29, 30] is based upon the approximation of sets by a pair of sets known as the lower approximation and the upper approximation of a set. Here, the lower and upper approximation operators are based on equivalence relation. However, the requirement of equivalence relation is a restrictive condition that may limit the application of the rough set model. Therefore, the basic notion of rough set is generalized in many ways. For instance, the equivalence relation is generalized to binary relations [15, 16, 24, 26, 27, 32], neighborhood systems [23], coverings [25], Boolean algebras [12, 31], fuzzy lattices [10], completely distributive lattices [5].

Further, the indiscernibility relation is generalized to almost indiscernibility relation to study many real life problems. The concept of rough set on fuzzy approximation spaces based on fuzzy proximity relation is studied by Tripathy and Acharjya [3, 7]. Further it is generalized to intuitionistic fuzzy proximity relation, and the concept of rough set on intuitionistic fuzzy approximation space is studied by Tripathy [1]. The different applications are also studied by Tripathy and Acharjya [2, 8, 9]. Further rough set of Pawlak is generalized to rough set on two universal sets with generalized approximation spaces [11, 20]. We continue a further study in the same direction.

One of our primary objectives in this article is to extend the results of Busse [13] in the context of rough set on two universal sets. Also, we establish necessary and sufficient type theorems and show how the results of Busse can be derived from them. The rest of the paper is organized as follows: Section 2 presents the foundations of rough set based on two universal sets. In Section 3, we study the approximation of classifications in the context of rough sets on two universal sets. The measures of uncertainty are studied in section 4. This is followed by a conclusion in Section 5.

## 2. Rough Set Based on Two Universal Sets

An information system is a table that provides a convenient way to describe a finite set of objects called the universe by a finite set of attributes thereby representing all available information and knowledge. But, in many real life situations, an information system establishes the relation between different universes. This gave the extension of rough set on single universal set to rough set on two universal sets. Wong et al. [20] generalized the rough set models using two distinct but related universal sets. Let $U$ and $V$ be two universal sets and $R \subseteq (U \times V)$ be a binary relation. By a knowledge base, we understand the relational system $(U, V, R)$ an approximation space. For an element $x \in U$, we define the right neighborhood or the $R$-relative set of $x$ in $U$, $r(x)$ as $r(x) = \cup \{y \in V : (x, y) \in R\}$. Similarly for an element $y \in V$, we define the left neighborhood or the $R$-relative set of $y$ in $V$, $l(y)$ as $l(y) = \cup \{x \in U : (x, y) \in R\}$.

For any two elements $x_1, x_2 \in U$, we say $x_1$ and $x_2$ are equivalent if $r(x_1) = r(x_2)$. Therefore, $(x_1, x_2) \in E_U$ if and only if $r(x_1) = r(x_2)$, where $E_U$ denote the equivalence relation on $U$. Hence, $E_U$ partitions the universal set $U$ into disjoint subsets. Similarly for any two elements $y_1, y_2 \in V$, we say $y_1$ and $y_2$ are equivalent if $l(y_1) = l(y_2)$. Thus, $(y_1, y_2) \in E_V$ if and only if $l(y_1) = l(y_2)$, where $E_V$ denote the equivalence relation on $V$ and partitions the universal set $V$ into disjoint subsets. Therefore for the approximation space $(U, V, R)$, it is clear that $E_V \circ R = R = R \circ E_U$, where $E_V \circ R$ is the composition of $R$ and $E_V$.

For any $Y \subseteq V$ and the binary relation $R$, we associate two subsets $\underline{R}Y$ and $\overline{R}Y$ called the $R$-lower and $R$-upper approximations of $Y$ respectively, which are given by:

$$\underline{R}Y = \bigcup \{x \in U : r(x) \subseteq Y\} \text{ and} \tag{1}$$

$$\overline{R}Y = \bigcup \{x \in U : r(x) \cap Y \neq \phi\}. \tag{2}$$

The $R$-boundary of $Y$ is denoted as $BN_R(Y)$ and is given as $BN_R(Y) = \overline{R}Y - \underline{R}Y$. The pair $(\underline{R}Y, \overline{R}Y)$ is called as the rough set of $Y \subseteq V$ if $\underline{R}Y \neq \overline{R}Y$ or equivalently $BN_R(Y) \neq \phi$.

Further, if $U$ and $V$ are finite sets, then the binary relation $R$ from $U$ to $V$ can be represented as $R(x, y)$, where

$$R(x, y) = \begin{cases} 1 & \text{if } (x, y) \in R \\ 0 & \text{if } (x, y) \notin R \end{cases}$$

The characteristic function of $X \subseteq U$ is defined for each $x \in U$ as follows:

$$X(x) = \begin{cases} 1 & \text{if } x \in X \\ 0 & \text{if } x \notin X \end{cases}$$

Therefore, the $R$- lower and $R$- upper approximations can also be presented in an equivalent form as shown below. Here $\wedge$ and $\vee$ denote the minimum and the maximum operators respectively.

$$(\underline{R}Y)x = \wedge_{y \in V} ((1 - R(x,y)) \vee Y(y)) \qquad (3)$$

and $\qquad (\overline{R}Y)x = \vee_{y \in V} (R(x,y) \wedge Y(y)) \qquad (4)$

**Definition 2.1** Let $U$ and $V$ be two universal sets. Let $R$ be a binary relation from $U$ to $V$. If $x \in U$ and $r(x) = \phi$, then we call $x$ is a solitary element with respect to $R$. The set of all solitary elements with respect to the relation $R$ is called as solitary set and is denoted as $S$. Mathematically,

$$S = \{x \in U : r(x) = \phi\} \qquad (5)$$

**2.1. Algebraic Properties of Rough Set based on Two Universal Sets**

In this section, we list the algebraic properties as established by Guilong Liu [11] those are interesting and valuable in the theory of rough sets as below. Let $R$ be an arbitrary binary relation from $U$ to $V$. Let $S$ be a solitary set with respect to the relation $R$. For subsets X, Y, in V

(i) $\overline{R}Y = \bigcup_{y \in Y} l(y)$

(ii) $\underline{R}\phi = S, \overline{R}\phi = \phi, \underline{R}V = U$ and $\overline{R}V = S'$, where $S'$ denotes the complement of $S$ in $U$.

(iii) $S \subseteq \underline{R}X$ and $\overline{R}X \subseteq S'$

(iv) $\underline{R}X - S \subseteq \overline{R}X$

(v) $\underline{R}X = U$ if and only if $\bigcup_{x \in U} r(x) \subseteq X$; $\overline{R}X = \phi$ if and only if $X \subseteq (\bigcup_{x \in U} r(x))'$

(vi) If $S \neq \phi$, then $\underline{R}X \neq \overline{R}X$ for all $X \in P(V)$, where $P(V)$ denotes the power set of V.

(vii) For any given index set $I$, $X_i \in P(V)$, $\underline{R}(\bigcap_{i \in I} X_i) = \bigcap_{i \in I} \underline{R}X_i$ and $\overline{R}(\bigcup_{i \in I} X_i) = \bigcup_{i \in I} \overline{R}X_i$.

(viii) If $X \subseteq Y$, then $\underline{R}X \subseteq \underline{R}Y$ and $\overline{R}X \subseteq \overline{R}Y$.

(ix) $\underline{R}X \cup \underline{R}Y \subseteq \underline{R}(X \cup Y)$, and $\overline{R}(X \cap Y) \subseteq \overline{R}X \cap \overline{R}Y$.

(x) $(\underline{R}X)' = \overline{R}X'$, and $(\overline{R}X)' = \underline{R}X'$;

(xi) There exists some $X \in P(U)$ such that $\underline{R}X = \overline{R}X$ iff R is serial.

79

(*xii*) If $G$ is another binary relation from $U$ to $V$ and $\overline{R}X = \overline{G}X$ for all $x \in P(V)$, then $R = G$.

(*xiii*) If $G$ is another binary relation from $U$ to $V$ and $\underline{R}X = \underline{G}X$ for all $x \in P(V)$, then $R = G$.

### 2.2. Characterization of Rough Set based on Two Universal Sets

In this section, we state an interesting topological characterization of rough set on two universal sets employing the notion of the lower and upper approximations as introduced by Tripathy and Acharjya [4]. It results in four important and different types of rough sets on two universal sets as shown below. Also, we discuss the set theoretic operations such as union and intersection on types of rough sets on two universal sets. However, for completeness of the paper we state the corresponding tables for union and intersection operations.

Type 1: If $\underline{R}Y \neq \phi$ and $\overline{R}Y \neq U$, then we say that Y is roughly R-definable on two universal sets.

Type 2: If $\underline{R}Y = \phi$ and $\overline{R}Y \neq U$, then we say that Y is internally R-undefinable on two universal sets.

Type 3: If $\underline{R}Y \neq \phi$ and $\overline{R}Y = U$, then we say that Y is externally R-undefinable on two universal sets.

Type 4: If $\underline{R}Y = \phi$ and $\overline{R}Y = U$, then we say that Y is totally R-undefinable on two universal sets.

**2.2.1. Table of union:** In the case of union as shown in Table 1, out of sixteen cases as many as nine cases are unambiguous whereas seven cases consist of ambiguity. In one case it can be in any one of the four types. These ambiguities are due to the inclusion $\underline{R}X \cup \underline{R}Y \subseteq \underline{R}(X \cup Y)$.

**Table 1. Table of Union**

| ∪ | Type 1 | Type 2 | Type 3 | Type 4 |
|---|---|---|---|---|
| **Type 1** | Type 1 / Type 3 | Type 1 / Type 3 | Type 3 | Type 3 |
| **Type 2** | Type 1 / Type 3 | Type 1 / Type 2 / Type 3 / Type 4 | Type 3 | Type 3 / Type 4 |
| **Type 3** | Type 3 | Type 3 | Type 3 | Type 3 |
| **Type 4** | Type 3 | Type 3 / Type 4 | Type 3 | Type 3 / Type 4 |

**2.2.2. Table of intersection:** It is interesting to see from the given Table 2 that, out of sixteen cases for intersection, seven cases are ambiguous whereas nine cases are unambiguous. Also, it is observed that, in one case it can be any one of the four types. These ambiguities are due to the inclusion $\overline{R}(X \cap Y) \subseteq \overline{R}X \cap \overline{R}Y$.

**Table 2. Table of Intersection**

| $\cap$ | Type 1 | Type 2 | Type 3 | Type 4 |
|---|---|---|---|---|
| **Type 1** | Type 1 / Type 2 | Type 2 | Type 1 / Type 2 | Type 2 |
| **Type 2** | Type 2 | Type 2 | Type 2 | Type 2 |
| **Type 3** | Type 1 / Type 2 | Type 2 | Type 1 / Type 2 / Type 3 / Type 4 | Type 2 / Type 4 |
| **Type 4** | Type 2 | Type 2 | Type 2 / Type 4 | Type 2 / Type 4 |

## 3. Approximation of Classifications

The rough set [29, 30] philosophy specifies about the understanding of the objects and their attributes influencing the objects with a depicted value. So, there is a need to classify objects of the universe based on the indiscernibility relation between them. The basic idea of rough set is based upon the approximation of sets by a pair of sets known as the lower approximation and upper approximation of the set. Here, the lower and upper approximation operators are based on equivalence relation. However, the requirement of equivalence relation is a restrictive condition that may limit the application of rough set model. Therefore, rough set is generalized by Guilong Liu [11] to rough set on two universal sets. Because we are interested in classifications based on binary relation, it is interesting to have the idea of approximation of classifications. It is because classifications of universes play central roles in rough set theory. Now, we define below a classification formally.

**Definition 3.1** Let $F = \{Y_1, Y_2, \cdots, Y_n\}$, where $n > 1$ be a family of non empty sets defined over $V$. We say that $F$ is a classification of $V$ if and only if $(Y_i \cap Y_j) = \phi$ for $i \neq j$ and $\sum_{k=1}^{n} Y_k = V$.

**Definition 3.2** Let $F = \{Y_1, Y_2, \cdots, Y_n\}$ be a family of non empty classification of $V$ and let $R$ be a binary relation from $U \to V$. Then the $R$-lower and $R$-upper approximation of the family $F$ is given as $\underline{R}F = \{\underline{R}Y_1, \underline{R}Y_2, \underline{R}Y_3, \cdots, \underline{R}Y_n\}$ and $\overline{R}F = \{\overline{R}Y_1, \overline{R}Y_2, \overline{R}Y_3, \cdots, \overline{R}Y_n\}$ respectively.

### 3.1. Theorems on Approximation of Classifications

In this section, we establish two theorems those are important in the context of knowledge representation, from which many corollaries can be derived including the four theorems established by Busse [13].

**Theorem 3.1** Let $R$ be a binary relation from $U \to V$ and let $F = \{Y_1, Y_2, \cdots, Y_n\}$, where $n > 1$ be a classification of $V$. For any $i \in \{1, 2, 3, \cdots, n\}$, $\overline{R}(\bigcup_i Y_i) = U$ if and only if $\underline{R}(\bigcup_j Y_j) = \phi$ for $j \neq i$ and $j \in \{1, 2, 3, \cdots, n\}$.

**Proof** If $\overline{R}(\bigcup_i Y_i) = U$, then for every $x \in U$ such that $r(x) \cap (\bigcup_i Y_i) \neq \phi$. This implies that $r(x) \subseteq Y_j$ does not hold for each $j \neq i$ and $j \in \{1, 2, 3, \cdots, n\}$. Therefore, $\underline{R}Y_j = \phi$ for all $j \neq i$ and $j \in \{1, 2, 3, \cdots, n\}$. Consequently $\underline{R}(\bigcup_j Y_j) = \phi$ for $j \neq i$ and $j \in \{1, 2, 3, \cdots, n\}$.

Conversely, if $\underline{R}(\bigcup_j Y_j) = \phi$ for $j \in \{1, 2, 3, \cdots, n\}$, then for each $x \in U$, $r(x) \subseteq Y_j$ does not hold for each $j \in \{1, 2, 3, \cdots, n\}$. It implies that for every $x \in U$, $r(x) \cap (\bigcup_i Y_i) \neq \phi$ for $i \neq j$ and $i \in \{1, 2, 3, \cdots, n\}$. Therefore, $\overline{R}(\bigcup_i Y_i) = U$.

**Corollary 3.1** Let $R$ be a binary relation from $U \to V$ and let $F = \{Y_1, Y_2, \cdots, Y_n\}$, where $n > 1$ be a classification of $V$. For any $i \in \{1, 2, 3, \cdots, n\}$, if $\overline{R}(\bigcup_i Y_i) = U$, then $\underline{R}Y_j = \phi$ for each $j \neq i$ and $j \in \{1, 2, 3, \cdots, n\}$.

**Proof** If $\overline{R}(\bigcup_i Y_i) = U$, then for every $x \in U$ such that $r(x) \cap (\bigcup_i Y_i) \neq \phi$. This implies that $r(x) \subseteq Y_j$ does not hold for each $j \neq i$ and $j \in \{1, 2, 3, \cdots, n\}$. Therefore, $\underline{R}Y_j = \phi$ for all $j \neq i$ and $j \in \{1, 2, 3, \cdots, n\}$.

**Corollary 3.2** Let $R$ be a binary relation from $U \to V$ and let $F = \{Y_1, Y_2, \cdots, Y_n\}$, where $n > 1$ be a classification of $V$. For each $i \in \{1, 2, 3, \cdots, n\}$, $\overline{R}Y_i = U$ if and only if $\underline{R}(\bigcup_j Y_j) = \phi$ for each $j \neq i$ and $j \in \{1, 2, 3, \cdots, n\}$.

**Proof** Taking only one value of $i$ in Theorem 3.1, we get this result.

**Corollary 3.3** Let $R$ be a binary relation from $U \to V$ and let $F = \{Y_1, Y_2, \cdots, Y_n\}$, where $n > 1$ be a classification of $V$. For each $i \in \{1, 2, 3, \cdots, n\}$, $\underline{R}Y_i = \phi$ if and only if $\overline{R}(\bigcup_j Y_j) = U$, for each $j \neq i$ and $j \in \{1, 2, 3, \cdots, n\}$.

**Proof** If $\underline{R}(Y_i) = \phi$ for each $i \in \{1, 2, 3, \cdots, n\}$, then for each $x \in U$, $r(x) \subseteq Y_i$ does not hold for each $i \in \{1, 2, 3, \cdots, n\}$. It implies that for every $x \in U$, $r(x) \cap (\bigcup_j Y_j) \neq \phi$ for $j \neq i$ and $j \in \{1, 2, 3, \cdots, n\}$. Therefore, $\overline{R}(\bigcup_j Y_j) = U$.

**Corollary 3.4 ([29] Proposition 2.6)** Let $R$ be a binary relation from $U \to V$ and let $F = \{Y_1, Y_2, \cdots, Y_n\}$, where $n > 1$ be a classification of $V$. If there exists $i \in \{1, 2, 3, \cdots, n\}$ such that $\overline{R}Y_i = U$, then for each $j \neq i$ and $j \in \{1, 2, 3, \cdots, n\}$ $\underline{R}Y_j = \phi$.

**Proof** If $\overline{R}Y_i = U$, then for every $x \in U$ such that $r(x) \cap Y_i \neq \phi$. This implies that $r(x) \subseteq Y_j$ does not hold for each $j \neq i$ and consequently $\underline{R}Y_j = \phi$ for each $j \neq i$ and $j \in \{1, 2, 3, \cdots, n\}$.

Similarly, $\underline{R}Y_1 = \{x_3\} \neq \phi$ with $\overline{R}Y_2 = \{x_2, x_4\} \neq U$ and $\overline{R}Y_3 = \{x_1, x_4, x_5\} \neq U$.

**Corollary 3.5 ([29] Proposition 2.8)** Let $R$ be a binary relation from $U \to V$ and let $F = \{Y_1, Y_2, \cdots, Y_n\}$, where $n > 1$ be a classification of $V$. If $\overline{R}Y_i = U$ for all $i \in \{1, 2, 3, \cdots, n\}$, then $\underline{R}Y_i = \phi$ for all $i \in \{1, 2, 3, \cdots, n\}$.

**Proof** Let us assume that $\overline{R}Y_i = U$ for all $i \in \{1, 2, 3, \cdots, n\}$. If for some $i$, $\underline{R}Y_i \neq \phi$, then there exists at least one $x \in U$ such that $r(x) \subseteq Y_i$. Since $(Y_i \cap Y_j) = \phi$, it implies that $r(x) \cap Y_j = \phi$ for $j \neq i$. Therefore, $\overline{R}Y_j \neq U$ for $j \neq i$ and $j \in \{1, 2, 3, \cdots, n\}$. Hence it is a contradiction. This completes the proof.

**Example 3.1** Let $U = \{x_1, x_2, x_3, x_4, x_5\}$ and $V = \{y_1, y_2, y_3, y_4, y_5, y_6\}$. Consider the relation $R$ given by its Boolean matrix as defined below.

$$R = \begin{pmatrix} 1 & 1 & 0 & 0 & 1 & 0 \\ 0 & 0 & 1 & 0 & 0 & 1 \\ 0 & 1 & 0 & 1 & 0 & 0 \\ 1 & 0 & 1 & 1 & 1 & 1 \\ 1 & 1 & 0 & 0 & 1 & 0 \end{pmatrix}$$

From the above relation $R$ it is clear that, $r(x_1) = \{y_1, y_2, y_5\}$; $r(x_2) = \{y_3, y_6\}$; $r(x_3) = \{y_2, y_4\}$; $r(x_4) = \{y_1, y_3, y_4, y_5, y_6\}$ and $r(x_5) = \{y_1, y_2, y_5\}$. Therefore, we get $U/E_U = \{\{x_1, x_5\}, \{x_2\}, \{x_3\}, \{x_4\}\}$. Similarly, $V/E_V = \{\{y_1, y_5\}, \{y_3, y_6\}, \{y_2\}, \{y_4\}\}$. Let the classification $C = \{Y_1, Y_2\}$ be given, where $Y_1 = \{y_1, y_2, y_6\}$ and $Y_2 = \{y_3, y_4, y_5\}$. Because $\overline{R}Y_1 = U = \overline{R}Y_2$ then $\underline{R}Y_1 = \phi$ and $\underline{R}Y_2 = \phi$.

Let the classification $C = \{Y_1, Y_2, Y_3\}$ be given, where $Y_1 = \{y_2, y_3, y_5\}$; $Y_2 = \{y_1, y_4\}$ and $Y_3 = \{y_6\}$. Because $\overline{R}Y_1 = \{x_1, x_2, x_3, x_4, x_5\} = U$, then $\underline{R}Y_2 = \phi$ and $\underline{R}Y_3 = \phi$. This verifies corollary 3.4. Similarly, the other corollaries mentioned above can also be verified through examples by taking different classification $C$.

**Theorem 3.2** Let $R$ be a binary relation from $U \to V$ and let $F = \{Y_1, Y_2, \cdots, Y_n\}$, where $n > 1$ be a classification of $V$. For any $i \in \{1, 2, 3, \cdots, n\}$, $\underline{R}(\bigcup_i Y_i) \neq \phi$ if and only if $\bigcup_j \overline{R}Y_j \neq U$ for $j \neq i$ and $j \in \{1, 2, 3, \cdots, n\}$.

**Proof** (*Necessity*) Suppose that $\underline{R}(\bigcup_i Y_i) \neq \phi$. Then there exists $x \in U$ such that $r(x) \subseteq (\bigcup_i Y_i)$. This implies that $r(x) \cap Y_j = \phi$ for all $j \neq i$ and $j \in \{1, 2, 3, \cdots, n\}$. It indicates that $x \notin \overline{R}Y_j$ for all $j \neq i$ and $j \in \{1, 2, 3, \cdots, n\}$. Consequently, $\bigcup_j \overline{R}Y_j \neq U$ for $j \neq i$ and $j \in \{1, 2, 3, \cdots, n\}$.

(*Sufficiency*) Let $i \in \{1, 2, 3, \cdots, n\}$. Suppose that $\bigcup_j \overline{R}Y_j \neq U$ for $j \neq i$ and $j \in \{1, 2, 3, \cdots, n\}$. By property of upper approximation, we have $\overline{R}\left(\bigcup_j Y_j\right) = \bigcup_j \overline{R}Y_j \neq U$. So there exists $r(x)$ for some $x \in U$ such that $r(x) \cap (\bigcup_j Y_j) = \phi$. It indicates that $r(x) \subseteq (\bigcup_i Y_i)$. Consequently, $\underline{R}(\bigcup_i Y_i) \neq \phi$.

**Corollary 3.6** Let $R$ be a binary relation from $U \to V$ and let $F = \{Y_1, Y_2, \cdots, Y_n\}$, where $n > 1$ be a classification of $V$. For any $i \in \{1, 2, 3, \cdots, n\}$, if $\underline{R}(\bigcup_i Y_i) \neq \phi$ then $\overline{R}Y_j \neq U$ for each $j \neq i$ and $j \in \{1, 2, 3, \cdots, n\}$.

**Proof** By Theorem 3.2, we have $\underline{R}(\bigcup_i Y_i) \neq \phi \Rightarrow \bigcup_j \overline{R}Y_j \neq U$. But by the property of upper approximation of union of rough set on two universal sets, we have $\bigcup_j \overline{R}Y_j \supseteq \overline{R}Y_j$ for each $j \neq i$ and $j \in \{1, 2, 3, \cdots, n\}$. Therefore, $\overline{R}Y_j \neq U$ for each $j \neq i$ and $j \in \{1, 2, 3, \cdots, n\}$.

**Corollary 3.7** Let $R$ be a binary relation from $U \to V$ and let $F = \{Y_1, Y_2, \cdots, Y_n\}$, where $n > 1$ be a classification of $V$. For any $i \in \{1, 2, 3, \cdots, n\}$, $\underline{R}Y_i \neq \phi$ if and only if $\bigcup_j \overline{R}Y_j \neq U$ for $j \neq i$ and $j \in \{1, 2, 3, \cdots, n\}$.

**Proof** On taking only one value for $i \in \{1, 2, 3, \cdots, n\}$ in Theorem 3.2 we get the desired result.

**Corollary 3.8** Let $R$ be a binary relation from $U \to V$ and let $F = \{Y_1, Y_2, \cdots, Y_n\}$, where $n > 1$ be a classification of $V$. For all $i$, $i \in \{1, 2, 3, \cdots, n\}$, $\overline{R}Y_i \neq U$ if and only if $\underline{R}(\bigcup_j Y_j) \neq \phi$ for $j \neq i$ and $j \in \{1, 2, 3, \cdots, n\}$.

**Proof** By Theorem 3.2, on taking the complement of indices we have $\bigcup_i \overline{R}Y_i \neq U \Rightarrow \underline{R}(\bigcup_j Y_j) \neq \phi$ for $j \neq i$ and $j \in \{1, 2, 3, \cdots, n\}$. On taking only one value for $i \in \{1, 2, 3, \cdots, n\}$ we get, for all $i$, $i \in \{1, 2, 3, \cdots, n\}$, $\overline{R}Y_i \neq U$ if and only if $\underline{R}(\bigcup_j Y_j) \neq \phi$ for $j \neq i$ and $j \in \{1, 2, 3, \cdots, n\}$.

**Corollary 3.9 ([29] Proposition 2.5)** Let $R$ be a binary relation from $U \to V$ and let $F = \{Y_1, Y_2, \cdots, Y_n\}$, where $n > 1$ be a classification of $V$. If there exists $i \in \{1, 2, 3, \cdots, n\}$ such that $\underline{R}Y_i \neq \phi$, then $\overline{R}Y_j \neq U$ for each $j \neq i$ and $j \in \{1, 2, 3, \cdots, n\}$.

**Proof** If $\underline{R}Y_i \neq \phi$, then there exists $x \in U$ such that $r(x) \subseteq Y_i$. This implies that $(r(x) \cap Y_j) = \phi$ for $j \neq i$ and $j \in \{1, 2, 3, \cdots, n\}$. Therefore, $\overline{R}Y_j \cap r(x) = \phi$ and consequently $\overline{R}Y_j \neq U$ for $j \neq i$ and $j \in \{1, 2, 3, \cdots, n\}$.

**Corollary 3.10 ([29] Proposition 2.7)** Let $R$ be a binary relation from $U \to V$ and let $F = \{Y_1, Y_2, \cdots, Y_n\}$, where $n > 1$ be a classification of $V$. If for all $i \in \{1, 2, 3, \cdots, n\}$, $\underline{R}Y_i \neq \phi$ holds then $\overline{R}Y_j \neq U$ for all $i \in \{1, 2, 3, \cdots, n\}$.

**Proof** As $\underline{R}Y_i \neq \phi$ for all $i \in \{1, 2, 3, \cdots, n\}$, we have $\underline{R}(\bigcup_i Y_i) \neq \phi$. Hence, by Corollary 3.8 $\overline{R}Y_i \neq U$ for all $i \in \{1, 2, 3, \cdots, n\}$. This completes the proof.

**Example 3.2** Let $U = \{x_1, x_2, x_3, x_4, x_5\}$ and $V = \{y_1, y_2, y_3, y_4, y_5, y_6\}$. Consider the relation $R$ given by its Boolean matrix:

$$R = \begin{pmatrix} 1 & 1 & 0 & 0 & 1 & 0 \\ 0 & 0 & 1 & 0 & 0 & 1 \\ 0 & 1 & 0 & 1 & 0 & 0 \\ 1 & 0 & 1 & 1 & 1 & 1 \\ 1 & 1 & 0 & 0 & 1 & 0 \end{pmatrix}$$

From the above relation $R$ it is clear that, $r(x_1) = \{y_1, y_2, y_5\}$; $r(x_2) = \{y_3, y_6\}$; $r(x_3) = \{y_2, y_4\}$; $r(x_4) = \{y_1, y_3, y_4, y_5, y_6\}$ and $r(x_5) = \{y_1, y_2, y_5\}$. Therefore, we get $U/E_U = \{\{x_1, x_5\}, \{x_2\}, \{x_3\}, \{x_4\}\}$. Similarly, $V/E_V = \{\{y_1, y_5\}, \{y_3, y_6\}, \{y_2\}, \{y_4\}\}$. Let us consider the classification $C = \{Y_1, Y_2\}$ where $Y_1 = \{y_1, y_2, y_4\}$ and $Y_2 = \{y_3, y_5, y_6\}$. Because $\underline{R}Y_1 = \{x_3\} \neq \phi$, $\underline{R}Y_2 = \{x_2\} \neq \phi$, then $\overline{R}Y_1 = \{x_1, x_3, x_4, x_5\} \neq U$ and $\overline{R}Y_2 = \{x_1, x_2, x_4, x_5\} \neq U$. This verifies Corollary 3.10.

Let the classification $C = \{Y_1, Y_2, Y_3\}$ be given, where $Y_1 = \{y_1, y_2, y_4\}$; $Y_2 = \{y_3, y_6\}$ and $Y_3 = \{y_5\}$. Because $\underline{R}Y_2 = \{x_2\} \neq \phi$, then $\overline{R}Y_1 = \{x_1, x_3, x_4, x_5\} \neq U$ and $\overline{R}Y_3 = \{x_1, x_4, x_5\} \neq U$. Similarly, $\underline{R}Y_1 = \{x_3\} \neq \phi$ with $\overline{R}Y_2 = \{x_2, x_4\} \neq U$ and $\overline{R}Y_3 = \{x_1, x_4, x_5\} \neq U$. This verifies the Corollary 3.9. Similarly, the other corollaries of Theorem 3.2 mentioned above can also be verified through examples by taking different classification $C$.

## 4. Measures of Uncertainty

In this section, we shall establish some properties of measures of uncertainty such as accuracy and quality of approximation employing the binary relation $R$ and discuss on properties of classifications. We denote the number of elements in a set $V$ by $card\ (V)$. Also, we establish three important theorems that are important in knowledge representation. Let $F = \{Y_1, Y_2, \cdots, Y_n\}$ be a family of non empty classifications. Then the $R$-lower and $R$-upper approximation of the family $F$ are given as $\underline{R}F = \{\underline{R}Y_1, \underline{R}Y_2, \underline{R}Y_3, \cdots, \underline{R}Y_n\}$ and $\overline{R}F = \{\overline{R}Y_1, \overline{R}Y_2, \overline{R}Y_3, \cdots, \overline{R}Y_n\}$ respectively. Now we define accuracy of approximation and quality of approximation of the family *F* employing the binary relation *R* as follows:

**Definition 4.1** The accuracy of approximation of *F* that expresses the percentage of possible correct decisions when classifying objects employing the binary relation *R* is defined as

$$\alpha_R(F) = \frac{\sum card\,(\underline{R}Y_i)}{\sum card\,(\overline{R}Y_i)} \text{ for } i = 1, 2, 3, \cdots, n \quad (6)$$

**Definition 4.2** The quality of approximation of *F* that expresses the percentage of objects which can be correctly classified to classes of *F* by the binary relation *R* is defined as

$$\nu_R(F) = \frac{\sum card\,(\underline{R}Y_i)}{card\,(V)} \text{ for } i = 1, 2, 3, \cdots, n \quad (7)$$

**Definition 4.3** We say that $F = \{Y_1, Y_2, \cdots, Y_n\}$ is *R*-definable if and only if $\underline{R}F = \overline{R}F$; that is $\underline{R}Y_i = \overline{R}Y_i$ for $i = 1, 2, 3, \cdots, n$.

**Theorem 4.1** Let *R* be a binary relation from $U \to V$ and let $F = \{Y_1, Y_2, \cdots, Y_n\}$, where $n > 1$ be a classification of *V*. For any *R*-definable classification *F* in *U*, $\alpha_R(F) = \nu_R(F) = 1$. Hence, if a classification *F* is *R*-definable then it is totally independent on *R*.

**Proof** For any *R*-definable classification *F*, $\underline{R}F = \overline{R}F$; that is $\underline{R}Y_i = \overline{R}Y_i$ for $i = 1, 2, 3, \cdots, n$. Therefore, by definition

$$\alpha_R(F) = \frac{\sum card\,(\underline{R}Y_i)}{\sum card\,(\overline{R}Y_i)} = 1$$

Again, by property of upper and lower approximation and as *F* is a classification of *V*, we have

$$\sum_{i=1}^{n} card\,(\overline{R}Y_i) \geq \sum_{i=1}^{n} card\,(Y_i) = card\left(\bigcup_{i=1}^{n} Y_i\right) = card\,(V) \text{ and}$$

$$\sum_{i=1}^{n} card\,(\underline{R}Y_i) \leq \sum_{i=1}^{n} card\,(Y_i) = card\left(\bigcup_{i=1}^{n} Y_i\right) = card\,(V)$$

But, for *R*-definable classifications, $\sum_{i=1}^{n} card\,(\underline{R}Y_i) = \sum_{i=1}^{n} card\,(\overline{R}Y_i)$ and hence we have $\sum_{i=1}^{n} card\,(\underline{R}Y_i) = card\,(V)$. Therefore, by definition

$$\nu_R(F) = \frac{\sum card\,(\underline{R}Y_i)}{card\,(V)} = 1$$

**Theorem 4.2** Let $R$ be a binary relation from $U \to V$ and let $F = \{Y_1, Y_2, \cdots, Y_n\}$, where $n > 1$ be a classification of $V$. If $\alpha_R(F) = v_R(F) = 1$, then $F$ is $R$-definable $V$.

**Proof** If $\alpha_R(F) = 1$, then by definition $\sum_{i=1}^{n} card(\underline{R}Y_i) = \sum_{i=1}^{n} card(\overline{R}Y_i)$. Again by definition of lower and upper approximation $card(\underline{R}Y_i) \leq card(\overline{R}Y_i)$. It indicates that $\underline{R}Y_i = \overline{R}Y_i$ for $i = 1, 2, 3, \cdots, n$. Therefore, then $F$ is $R$-definable $V$.

**Theorem 4.3** Let $R$ be a binary relation from $U \to V$ and for any classification $F = \{Y_1, Y_2, \cdots, Y_n\}$, $n > 1$ in $V$, $0 \leq \alpha_R(F) \leq v_R(F) \leq 1$.

**Proof** By property of lower approximation and as $F$ is a classification of $V$, we have $\sum_{i=1}^{n} card(\underline{R}Y_i) \leq \sum_{i=1}^{n} card(Y_i) = card\left(\bigcup_{i=1}^{n} Y_i\right) = card(V)$. Therefore, by definition we have $v_R(F) = \dfrac{\sum card(\underline{R}Y_i)}{card(V)} \leq \dfrac{card(V)}{card(V)} = 1$.

Again, $\sum_{i=1}^{n} card(\overline{R}Y_i) \geq \sum_{i=1}^{n} card(Y_i) = card\left(\bigcup_{i=1}^{n} Y_i\right) = card(V)$. Hence, by definition we have $\alpha_R(F) = \dfrac{\sum card(\underline{R}Y_i)}{\sum card(\overline{R}Y_i)} \leq \dfrac{\sum card(\underline{R}Y_i)}{card(V)} = v_R(F)$ ; i.e., $\alpha_R(F) \leq v_R(F)$ and consequently $0 \leq \alpha_R(F) \leq v_R(F) \leq 1$.

## 5. Conclusion

In this paper we extended the study of rough set on two universal sets further by defining approximation of classifications in rough set on two universal sets. We considered the types of union and intersection of rough sets on two universal sets. Also, we generalized the four theorems established by Busse [13] on approximation of classifications and obtain two theorems of necessary and sufficient type to the settings of rough sets on two universal sets. From these theorems several other results besides Busse's theorems could be derived as corollaries in the settings of rough set on two universal sets. We have also defined the accuracy and quality of approximation of classifications on two universal sets. Also we describe through theorems the inexactness of classification using topological means as in the case of sets.

# Authors

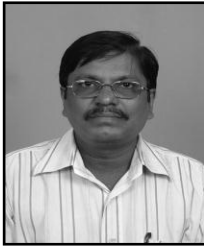

**B K Tripathy**, a senior professor in the school of computing sciences and engineering, VIT University, at Vellore, India. He has been awarded with Gold Medals both at graduate and post graduate levels of Berhampur University, India. Also, he has been awarded with the best post graduate of the Berhampur University. He has published more than 140 technical papers in various international journals, conferences, and Springer book chapters. He is associated with many professional bodies like IEEE, ACM, IRSS, WSEAS, AISTC, ISTP, CSI, AMS, and IMS. He is in the editorial board of several international journals like CTA, ITTA, AMMS, IJCTE, AISS, AIT, and IJPS. Also, he is a reviewer of international journals like Mathematical Reviews, Information Sciences, Analysis of Neural Networks, Journal of Knowledge Engineering, Mathematical Communications, IJET, IJPR and Journal of Analysis. His research interest includes fuzzy sets and systems, rough sets and knowledge engineering, data clustering, granular computing and social networks.

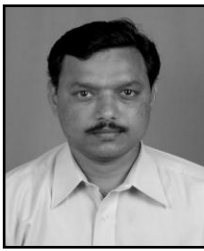

**D P Acharjya** received the M. Tech. degree in computer science from Utkal University, India in 2002; M. Phil. from Berhampur University, India; and M. Sc. from NIT, Rourkela, India. He has been awarded with Gold Medal in M. Sc. He is a research student of Berhampur University, India. Currently, he is an Associate Professor in the school of computing sciences and engineering, VIT University, Vellore, India. He has authored many national and international journal papers and three books to his credit. He is associated with many professional bodies CSI, ISTE, IMS, AMTI, ISIAM, OITS, IACSIT, CSTA, IEEE and IAENG. He was founder secretary of OITS Rourkela chapter. His current research interests include rough sets, formal concept analysis, knowledge representation, granular computing, data mining and business intelligence.